# The Chan-Vese Algorithm

## Project Report

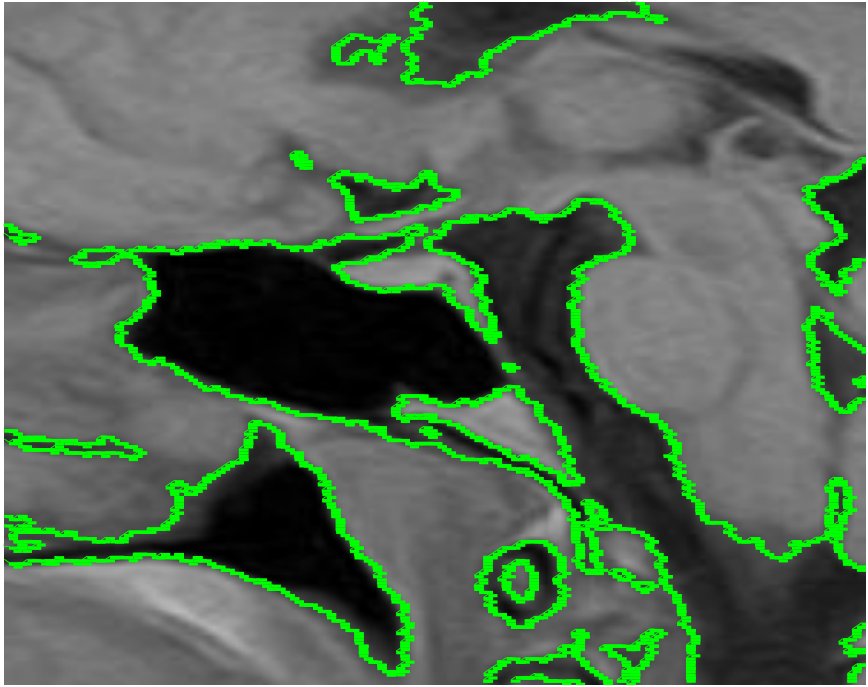

Rami Cohen, rc@tx.technion.ac.il

Introduction to Medical Imaging, Spring 2010

Technion, Israel Institute of Technology

This report is accompanied by a MATLAB package that can be requested by mail.

# Table of Contents

# 1. Introduction

Segmentation is the process of partitioning a digital image into multiple segments (sets of pixels). Such common segmentation tasks including segmenting written text or segmenting tumors from healthy brain tissue in an MRI image, etc.

Chan-Vese model for active contours [1] is a powerful and flexible method which is able to segment many types of images, including some that would be quite difficult to segment in means of "classical" segmentation – i.e., using thresholding or gradient based methods.

This model is based on the Mumford-Shah functional [2] for segmentation, and is used widely in the medical imaging field, especially for the segmentation of the brain, heart and trachea [3].

The model is based on an energy minimization problem, which can be reformulated in the level set formulation, leading to an easier way to solve the problem.

In this project, the model will be presented (there is an extension to color (vector-valued) images [4], but it will not be considered here), and MATLAB code that implements it will be introduced.

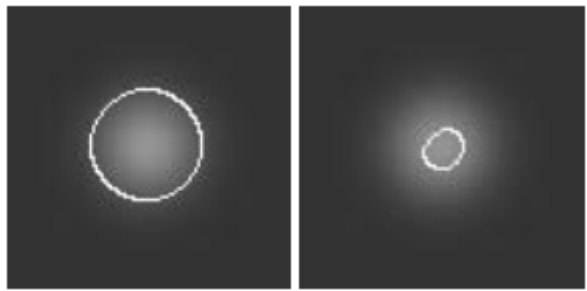

Figure 1: left: using Chan-Vese method, right: using gradient based method

## 2. The model

Let $\Omega$ be a bounded open set of $\mathbb{R}^2$, with $\partial\Omega$ its boundary. Let $u_0 : \bar{\Omega} \to \mathbb{R}$ be a given image, and $C(s)$ is a piecewise $C^1[0,1]$ parameterized a curve.

Let's denote the region inside $C$ as $\omega$, and the region outside $C$ as $\bar{\Omega}\setminus\omega$. Moreover, $c_1$ will denote the average pixels' intensity inside $C$, and $c_2$ will denote the average intensity outside $C$ (i.e., $c_1 = c_1(C), c_2 = c_2(C)$).

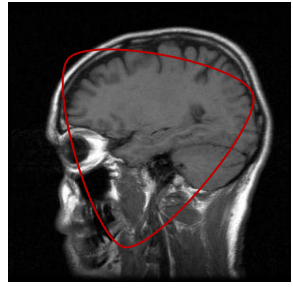

Figure 2: the image is definded on $\bar{\Omega}$ (the big rectangle), the (arbitrary) red curve is C

The object of Chan-Vese algorithm is to minimize the energy functional $F(c_1, c_2, C)$, defined by:

$$F(c_1, c_2, C) = \mu \cdot Length(C) + v \cdot Area(inside(C))$$
$$+ \lambda_1 \int_{inside(C)} |u_0(x,y) - c_1|^2 dxdy + \lambda_2 \int_{outside(C)} |u_0(x,y) - c_2|^2 dxdy$$

Equation 1: The energy functional

Where $\mu \geq 0, v \geq 0, \lambda_1, \lambda_2 > 0$ are fixed parameters (should be determined by the user). As suggested by the paper, the preferred settings are $v = 0, \lambda_1 = \lambda_2 = 1$. It should be noted that the term $Length(C)$ could be re-written more generally as $(Length(C))^p$ for $p \geq 1$, but usually $p = 1$.

In other words, we are looking for $c_1, c_2, C$ that will be the solution to the minimization problem:

$$\inf_{c_1, c_2, C} F(c_1, c_2, C)$$

Equation 2: The minimization problem

## 3. Level Set formulation

Instead of searching for the solution in terms of $C$, we can redefine the problem in the level set formalism. In the level set method, $C \subset \Omega$ is represented by the zero level set of some Lipschitz function[1] $\Phi : \Omega \to \mathbb{R}$, s.t.:

$$\begin{cases} C = \partial \omega = \{(x,y) \in \Omega : \Phi(x,y) = 0\} \\ inside(C) = \omega = \{(x,y) \in \Omega : \Phi(x,y) > 0\} \\ outside(C) = \Omega \setminus \bar{\omega} = \{(x,y) \in \Omega : \Phi(x,y) < 0\} \end{cases}$$

Equation 3: Level set formulation

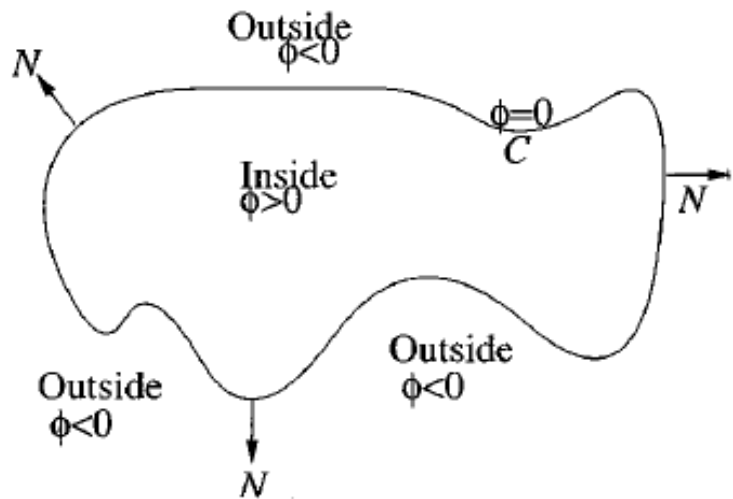

Figure 3: The signed distance function

In the following implementation, given a contour $C$, $\Phi(x,y)$ is defined as the signed distance function from $C$, where outside $C$ the sign of $\Phi(x,y)$ is negative.

---

[1] Lipschitz function is a function $f$ such that $|f(x) - f(y)| \leq C|x - y|$ for some constant $C$ which is independent of $x$ and $y$. For example, any function with a bounded first derivative is Lipschitz.

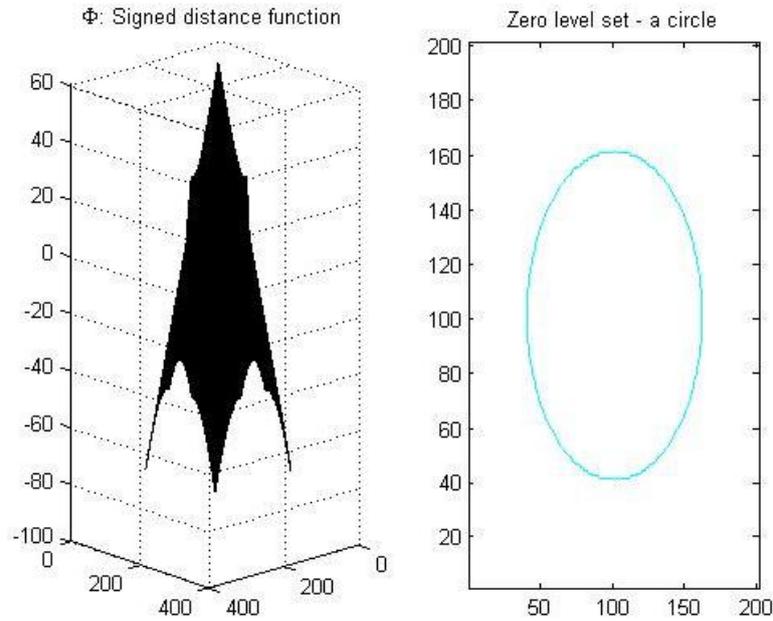

Figure 4: Example for $\Phi$ and its zero level set

Our object is to evolve $\Phi(x, y)$, when the evolved contour $C$ in each time $t$ is the zero level set of $\Phi(x, y, t)$.

We can re-write the functional $F(c_1, c_2, C)$ in terms of $\Phi(x, y)$ only:

1. $Length(C)$ can be calculated as the length of the zero level set $\Phi(x, y) = 0$:

$$Length(C) = \int_{\Omega} |\nabla H(\Phi(x, y))| \, dxdy = \int_{\Omega} \delta_0(\Phi(x, y)) |\nabla \Phi(x, y)| \, dxdy$$

Where $H(z)$ is the Heaviside function: $H(z) = \begin{cases} 1, & \text{if } z \geq 0 \\ 0, & \text{if } z < 0 \end{cases}$

2. $Area(inside(C))$ can be calculated as the area of the region in which $\Phi(x, y) \geq 0$:

$$Area(inside(C)) = \int_{\Omega} H(\Phi(x, y)) \, dxdy$$

3. $\int_{inside(C)} |u_0(x, y) - c_1|^2 \, dxdy$ can be calculated in terms of $\Phi(x, y)$, when considering only the region in which $\Phi(x, y) > 0$:

$$\int_{inside(C)} |u_0(x,y)-c_1|^2 dxdy = \int_{(x,y):\Phi(x,y)>0} |u_0(x,y)-c_1|^2 dxdy = \int_\Omega |u_0(x,y)-c_1|^2 H(\Phi(x,y)) dxdy$$

4. in a similar way:

$$\int_{outside(C)} |u_0(x,y)-c_2|^2 dxdy = \int_{(x,y):\Phi(x,y)<0} |u_0(x,y)-c_2|^2 dxdy = \int_\Omega |u_0(x,y)-c_2|^2 H(1-\Phi(x,y)) dxdy$$

5. The average intensities:

$$c_1 = \frac{\int_\Omega u_0(x,y) H(\Phi(x,y)) dxdy}{\int_\Omega H(\Phi(x,y)) dxdy}, \quad c_2 = \frac{\int_\Omega u_0(x,y) H(1-\Phi(x,y)) dxdy}{\int_\Omega H(1-\Phi(x,y)) dxdy}$$

The above leads us to the energy functional in terms of $(c_1, c_2, \Phi)$ (where $c_1 = c_1(\Phi), c_2 = c_2(\Phi)$ and $\delta_0(x)$ is Dirac delta function):

$$F(c_1, c_2, \Phi) = \mu \int_\Omega \delta_0(\Phi(x,y)) |\nabla \Phi(x,y)| dxdy + v \int_\Omega H(\Phi(x,y)) dxdy$$
$$+ \lambda_1 \int_\Omega |u_0(x,y)-c_1|^2 H(\Phi(x,y)) dxdy + \lambda_2 \int_\Omega |u_0(x,y)-c_2|^2 H(1-\Phi(x,y)) dxdy$$

**Equation 4: Energy functional in terms of $\Phi$**

Observing the terms in equation 4, we can say that the evolution of the curve is influenced by two terms ($v$ is usually set to 0, so we will ignore it): the curvature regularizes the curve and makes it smooth during evolution; the "region term" $-\lambda_1(u_0 - c_1)^2 + \lambda_2(u_0 - c_2)^2$ affects the motion of the curve [5].

The term $\mu \int_\Omega \delta_0(\Phi(x,y)) |\nabla \Phi(x,y)| dxdy$ is the penalty on the total length of the curve $C$. For example, if the boundaries of the image are quite smooth, we will give $\mu$ a larger value, to prevent $C$ from being a complex curve.

$\lambda_1, \lambda_2$ affect the desired uniformity inside $C$ and outside $C$, respectively. For example, It would be advisable to set $\lambda_1 < \lambda_2$ when we expect an image with quite uniform background and varying grayscale objects in the foreground.

Using Euler-Lagrange equations and the gradient-descent method, it is shown in the paper that $\Phi(x, y)$ which minimizes the energy $F(c_1, c_2, \Phi)$ satisfies the PDE ($t$ is an artificial time):

$$\begin{cases} \dfrac{\partial \Phi}{\partial t} = \delta(\Phi) \left[ \mu \cdot p \cdot \left( \int_\Omega \delta(\Phi) |\nabla \Phi| dxdy \right)^{p-1} \mathrm{div}\left( \dfrac{\nabla \Phi}{|\nabla \Phi|} \right) - v - \lambda_1(u_0 - c_1)^2 + \lambda_2(u_0 - c_2)^2 \right] \\ = \delta(\Phi) \left[ \mu \cdot p \cdot \left( \int_\Omega \delta(\Phi) |\nabla \Phi| dxdy \right)^{p-1} \cdot \kappa(\Phi) - v - \lambda_1(u_0 - c_1)^2 + \lambda_2(u_0 - c_2)^2 \right] \\ \Phi(x, y, 0) = \Phi_o(x, y), \quad \dfrac{\delta(\Phi)}{|\nabla \Phi|} \dfrac{\partial \Phi}{\partial \vec{n}} = 0 \text{ on } \partial \Omega \end{cases}$$

Equation 5: PDE for $\Phi(x, y, t)$

Where $\kappa(\Phi)$ is the *curvature* of the evolving curve (for some specific height level in $\Phi$). We saw at class that the curvature can be calculated using the spatial derivatives of $\Phi$ up to second order:

$$\kappa(\Phi) = \frac{\Phi_{xx} \Phi_y^2 - 2\Phi_{xy} \Phi_x \Phi_y + \Phi_{yy} \Phi_x^2}{\left( \Phi_x^2 + \Phi_y^2 \right)^{3/2}}$$

Equation 6: Curvature

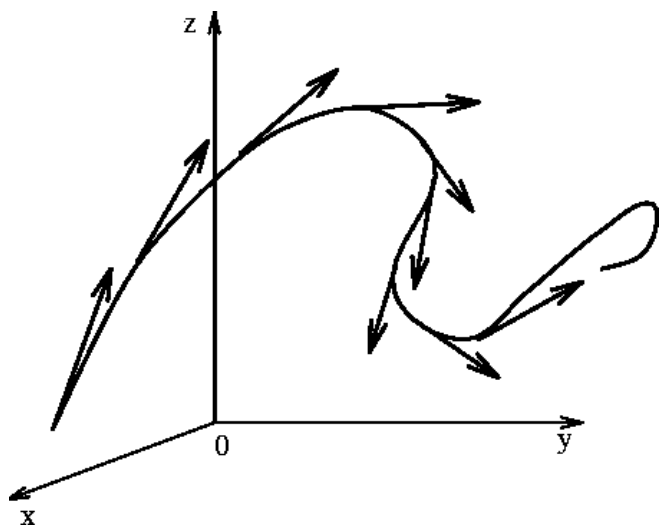

Figure 5: Curvature at a point can tell how fast the curvature is turning there

## 4. Numerical scheme

First, we define regularizations of $H(x)$ and $\delta(x)$ (where $\delta(x) = \dfrac{d}{dx}H(x)$):

$$H_\varepsilon(x) = \frac{1}{2}\left(1 + \frac{2}{\pi}\arctan\left(\frac{x}{\varepsilon}\right)\right)$$

$$\delta_\varepsilon(x) = \frac{1}{\pi}\frac{\varepsilon}{\varepsilon^2 + x^2}$$

**Equation 7: Regularizations of the Heaviside and Dirac delta function**

For some constant $\varepsilon > 0$. The values used in the simulations are $\varepsilon = h = 1$, where $h$ is the space step (it is reasonable to choose $\varepsilon = h$, since $h$ is the smallest space step in the problem). These regularizations achieve good results in simulations, as described in the paper, in the sense that they usually lead to the global minimum of the energy.

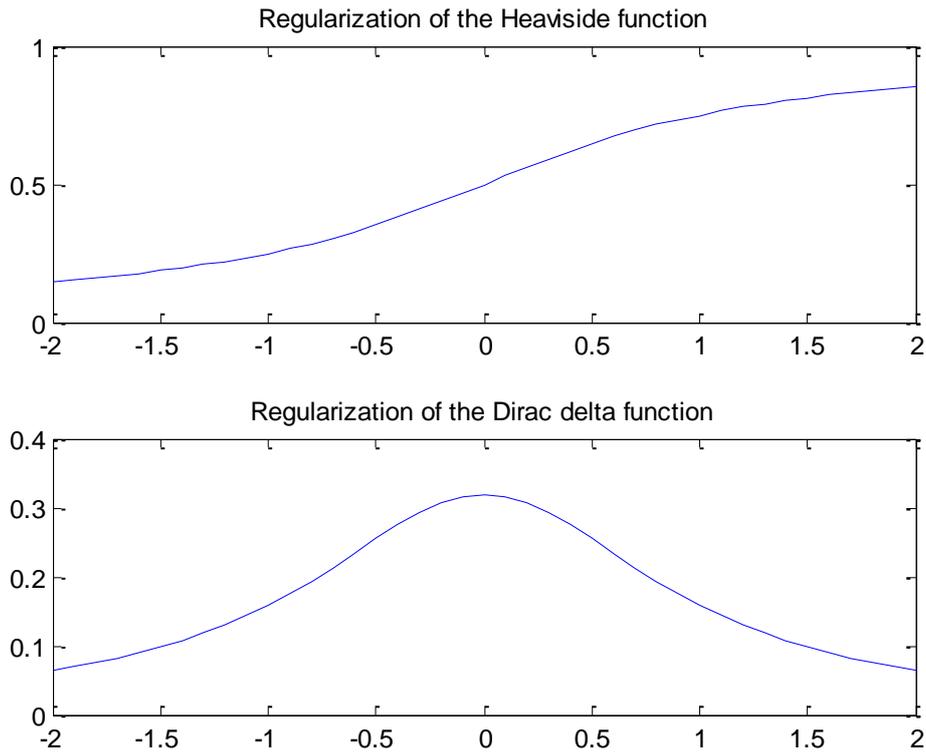

**Figure 6: Regularizations**

Let's define $\Phi_{i,j}^n = \Phi(n\Delta t, x_i, y_j)$ where $\Delta t$ is the time step. The PDE can be discretisized by using the following notations for spatial finite differences (where $h_x = h_y = h$):

$$\Delta_-^x \Phi_{i,j}^n = (\Phi_{i,j}^n - \Phi_{i-1,j}^n)/h, \quad \Delta_+^x \Phi_{i,j}^n = (\Phi_{i+1,j}^n - \Phi_{i,j}^n)/h$$

$$\Delta_-^y \Phi_{i,j}^n = (\Phi_{i,j}^n - \Phi_{i,j-1}^n)/h, \quad \Delta_+^y \Phi_{i,j}^n = (\Phi_{i,j+1}^n - \Phi_{i,j}^n)/h$$

The linearized, discretized PDE becomes:

$$\frac{\Phi_{i,j}^{n+1} - \Phi_{i,j}^n}{\Delta t} = \delta_h(\Phi_{i,j}^n) \frac{\mu}{h^2} \left[ \Delta_-^x \left( \frac{\Delta_+^x \Phi_{i,j}^n}{\sqrt{(\Delta_+^x \Phi_{i,j}^n)^2/h^2 + (\Phi_{i,j+1}^n - \Phi_{i,j-1}^n)^2/(2h)^2}} \right) + \Delta_-^y \left( \frac{\Delta_+^y \Phi_{i,j}^n}{\sqrt{(\Delta_+^y \Phi_{i,j}^n)^2/h^2 + (\Phi_{i+1,j}^n - \Phi_{i-1,j-1}^n)^2/(2h)^2}} \right) \right]$$

$$-\delta_h(\Phi_{i,j}^n)\left(v + \lambda_1 (u_{o_{i,j}} - c_1(\Phi^n))^2 - \lambda_2 (u_{o_{i,j}} - c_2(\Phi^n))^2 \right)$$

**Equation 8: The linearized, discretized PDE**

Defining the following constants (for a given $\Phi^n$):

$$C_1 = \frac{1}{\sqrt{(\Phi_{i+1,j}^n - \Phi_{i,j}^n)^2 + (\Phi_{i,j+1}^n - \Phi_{i,j-1}^n)^2/4}}, \quad C_2 = \frac{1}{\sqrt{(\Phi_{i,j}^n - \Phi_{i-1,j}^n)^2 + (\Phi_{i-1,j}^n - \Phi_{i-1,j-1}^n)^2/4}}$$

$$C_3 = \frac{1}{\sqrt{(\Phi_{i+1,j}^n - \Phi_{i-1,j}^n)^2/4 + (\Phi_{i,j+1}^n - \Phi_{i,j}^n)^2}}, \quad C_4 = \frac{1}{\sqrt{(\Phi_{i+1,j-1}^n - \Phi_{i-1,j-1}^n)^2/4 + (\Phi_{i,j}^n - \Phi_{i,j-1}^n)^2}}$$

We get the simplified equation:

$$\Phi_{i,j}^{n+1} \left[ 1 + \frac{\Delta t}{h} \delta_h(\Phi_{i,j}^n) \cdot \mu \cdot \left( p \cdot L(\Phi^n)^{p-1} \right)(C_1 + C_2 + C_3 + C_4) \right]$$

$$= \Phi_{i,j}^n + \frac{\Delta t}{h} \delta_h(\Phi_{i,j}^n) \cdot \mu \cdot \left( p \cdot L(\Phi^n)^{p-1} \right)\left(C_1 \Phi_{i+1,j}^n + C_2 \Phi_{i-1,j}^n + C_3 \Phi_{i,j+1}^n + C_4 \Phi_{i,j-1}^n \right)$$

$$-\Delta t \delta_h(\Phi_{i,j}^n)\left[ v + \lambda_1 (u_{0_{i,j}} - c_1(\Phi^n))^2 - \lambda_2 (u_{0_{i,j}} - c_2(\Phi^n))^2 \right]$$

**Equation 9: Simplified PDE**

Where we calculate $c_1(\Phi^n), c_2(\Phi^n)$ as discretized sums, using the regularized Heaviside function. In addition, $L(\Phi^n)$ is $Length(C)$ as was calculated at section 3.

As suggested in the paper, in order to solve equation 10, we can use an iterative way, as pointed out by [6], proposition 6.1, in which it is also proved that there exists a solution to the equation.

### 4.1 Reinitialization of $\Phi$

In each step, we need to reinitialize $\Phi(x, y)$ to be the signed distance function to its zero level set. This procedure prevents the level set function from becoming too "flat", an effect which is caused due the use of the regularized delta function $\delta_\varepsilon(x)$, which causes blurring.

The reinitialization process is made by using the following PDE:

$$\begin{cases} \dfrac{\partial \Psi}{\partial \tau} = sign(\Phi(x,y,t))(1-|\nabla \Psi|) \\ \Psi(0) = \Phi(x,y,t) \end{cases}$$

**Equation 10: Reinitialization**

The solution of this equation, $\Psi$, will have the same zero level set as $\Phi(x, y, t)$, and away from this level set, $|\nabla \Psi|$ will converge to 1, as it should be for a distance function.

The numerical equation for equation 11:

$$\Psi_{i,j}^{n+1} = \Psi_{i,j}^n - \Delta\tau\, sign(\Phi(x,y,t)) G(\Psi_{i,j}^n)$$

**Equation 11: Numerical scheme for reinitialization**

Where the "flux" $G(\Psi_{i,j}^n)$ is defined using the notations $a, b, c, d$, defined by:

$$a = \left(\Delta_-^x \Psi_{i,j}\right)/h = \left(\Psi_{i,j} - \Psi_{i-1,j}\right)/h$$
$$b = \left(\Delta_+^x \Psi_{i,j}\right)/h = \left(\Psi_{i+1,j} - \Psi_{i,j}\right)/h$$
$$c = \left(\Delta_-^y \Psi_{i,j}\right)/h = \left(\Psi_{i,j} - \Psi_{i,j-1}\right)/h$$
$$d = \left(\Delta_+^y \Psi_{i,j}\right)/h = \left(\Psi_{i,j+1} - \Psi_{i,j}\right)/h$$

and

$$G\left(\Psi_{i,j}^n\right) = \begin{cases} \sqrt{\max\left(\left(a^+\right)^2,\left(b^-\right)^2\right) + \max\left(\left(c^+\right)^2,\left(d^-\right)^2\right)} - 1, & \Phi(x_i, y_j, t) > 0 \\ \sqrt{\max\left(\left(a^-\right)^2,\left(b^+\right)^2\right) + \max\left(\left(c^-\right)^2,\left(d^+\right)^2\right)} - 1, & \Phi(x_i, y_j, t) < 0 \\ 0, & \text{otherwise} \end{cases}$$

**Equation 12: definition of $G$**

where $a^+ = \max(a,0), a^- = \min(a,0)$ and so on.

### 4.2 Summary of the algorithm

1. Initialize $\Phi_{i,j}^{n=0}$ to some Lipschitz function $\Phi_0$
2. Compute $c_1(\Phi^n), c_2(\Phi^n)$
3. Solve the PDE of equation 9
4. Reinitialize $\Phi_{i,j}^{n+1}$ to be the signed distance function to $\{\Phi_{i,j}^{n+1} = 0\}$ by using equation 11
5. Check whether the solution is stationary. If not, continue. Else, stop.

In practice, the process should be stopped when $Q = \sum_{|\Phi_{i,j}^n| < h} \left|\Phi_{i,j}^{n+1} - \Phi_{i,j}^n\right| \leq \Delta t \cdot h^2$ (this should be checked at stage no. 5) where $M$ is the number of grid points which satisfy $|\Phi_{i,j} < h|$, because $\Phi(x, y, t)$ is not expected to change anymore (except for maybe some small numerical changes).

## 5. Code

The numerical scheme above was implemented using MATLAB. The attached source files are (more documentation can be found in the source code):

1. CV.m – the main file.
2. delta.m – regularized Dirac delta function (equation 7).
3. heaviside.m - regularized Heaviside function (equation 7).
4. reinit.m – reinitialization of $\Phi_{i,j}^{n+1}$ by the method proposed at 4.1.
5. cur_diff – checking whether $\Phi_{i,j}^{n}$ reached its steady state according to the method proposed at 4.2

The input images should be grayscale only; otherwise they will be converted into grayscale. Moreover, there is an option to select only a (rectangular) part of the image, and the segmentation process will be applied only to this part. It can come in handy when the image is large and we want to segment just a part of it.

As always, larger images need more computational time. It is highly recommended to use images of a small size (i.e., no more than 256x256 pixels), and to not stop the segmentation process before it ends (in other words, don't use ctrl+c). In weak computers it can cause some instability.

## 6. Results

The first example is similar to the example which appears at the paper – an image with (approx.) only two gray levels.

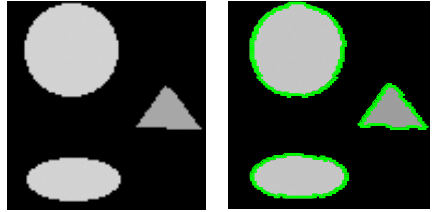

Figure 7: left: clear objects, right: segmentation

As can be seen from figure, a good segmentation was achieved.

Let's see how it segments an MRI image (resized):

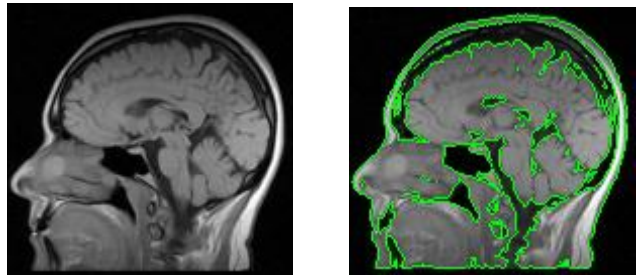

Figure 8: left: MRI image, right: segmentation

Since Chan-Vese algorithm is not based on gradient methods, it achieves good results even when the image is blurred:

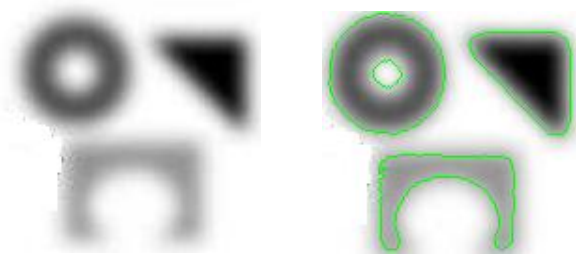

Figure 9: left: blurred objects, right: segmentation

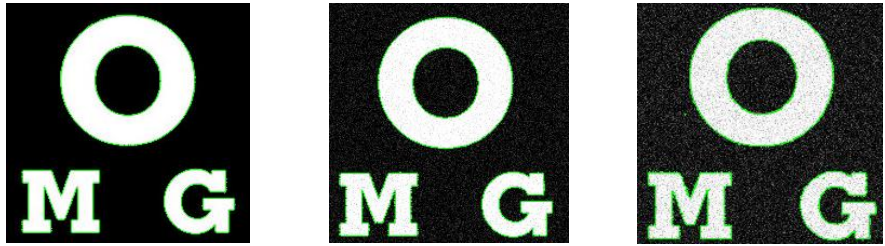

Figure 10: left: no noise, middle: modest noise, right: stronger noise

In figure 10 we can see that the algorithm deals quite well with noisy images ( $p$ was set to 2 in this case, to prevent $C$ from surrounding the noisy dots).

The algorithm can also detect quite precisely thin edges:

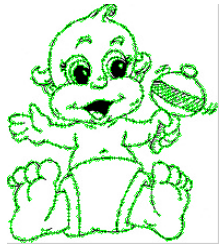

Figure 11: segmentaion of image with thin edges

Some optimization has been applied to the code, in order to decrease running time (working on matrices without loops, etc.). For example, Image of size $200 \times 200$ pixels needs on average about 0.25 minute (when using the maximal number of iterations) for the segmentation process. When preferring shorter time on accuracy, a quite fair segmentation can be achieved within just several seconds (3-10 iterations).

The complexity of the algorithm is $O(MN)$ (in each iteration) where $M \times N$ is the size of the image. Improvement for running time can be achieved by applying any efficient method that can detect the regions in $\Phi(x, y)$ which undergo the main changes, so the updating process will concentrate on these regions (see later).

Of course, many other results can be obtained by simply using the attached code package, with proper selection of the parameters. Using the default parameters would also give fair results.

## 7. Conclusions

Chan-Vese algorithm was implemented in this project. From the results above, it can be seen that this algorithm deals quite well even with images which are quite difficult to segment in the regular methods, such as gradient-based methods or thresholding.

This can be explained by the fact that CV algorithm relies on global properties (intensities, region areas), rather than just taking into account local properties, such as gradients. One of the main advantages of this approach is better robustness for noise, for example.

As mentioned before, the algorithm is sometimes quite slow, especially when dealing with large images. It can pose a problem for real time applications, such as video sequences, and an efficient implementation is very important.

There are some papers which suggest refinements to this algorithm, especially for the time-consuming computation of the PDE solution. These methods use values that were already computed, in order to decrease the computing time of the next values. Such approaches can be found in [5], [7] .

CV algorithm is a very powerful algorithm, as we have also seen in the results above. This algorithm marks some "modern" approach for image segmentation, which relies on calculus and partial differential equations.

# 9. Appendix

## 9.1 Table of equations

## 9.2 Table of figures